\documentclass[journal]{IEEEtran}
\usepackage{blindtext}
\usepackage{graphicx}
\usepackage{cite}
\usepackage{url}
\usepackage{amsmath}
\usepackage{arydshln} 
\usepackage{rotating}
\usepackage{pifont}
\usepackage{tikz}
\usetikzlibrary{fit,shapes,backgrounds,decorations.pathreplacing,calc}

\newcommand{\cmark}{\ding{51}}%
\newcommand{\xmark}{\ding{55}}%

\begin{document}
\title{Survey on Models and Techniques for Root-Cause Analysis}

\author{Marc~Sol\'e
        \hspace{5mm}
        Victor~Munt\'es-Mulero
        \thanks{M. Sol\'e and V. Munt\'es are with CA Technologies.}
        \hspace{5mm}
        Annie~Ibrahim Rana
        \hspace{5mm}
        Giovani~Estrada
        \thanks{A. Ibrahim and G. Estrada are with Intel}
        }


\IEEEpeerreviewmaketitle




\maketitle

\begin{abstract}
\boldmath
Automation and computer intelligence to support complex human decisions becomes essential to manage large and distributed systems in the Cloud and IoT era. Understanding the root cause of an observed symptom in a complex system has been a major problem for decades. As industry dives into the IoT world and the amount of data generated per year grows at an amazing speed, an important question is how to find appropriate mechanisms to determine root causes that can handle huge amounts of data or may provide valuable feedback in real-time. While many survey papers aim at summarizing the landscape of techniques for modelling system behavior and infering the root cause of a problem based in the resulting models, none of those focuses on analyzing how the different techniques in the literature fit growing requirements in terms of performance and scalability. In this survey, we provide a review of root-cause analysis, focusing on these particular aspects. We also provide guidance to choose the best root-cause analysis strategy depending on the requirements of a particular system and application. 
\end{abstract}


\section{Introduction}

With the onset of the SMAC industry (i.e. Social, Mobile, Analytics and Cloud), Software as a Service, and the Internet of Things (IoT), more organizations in all industry sectors recognize they are evolving into technology and data companies\footnote{McKendrick, J. (2015, April 30). Every Company Now A Technology Company: Latest Round Of Mergers And Acquisitions Confirms It. Forbes. Retrieved from http://tinyurl.com/j6f7ub5}. While one may be tempted to believe that this may only affect some limited number of markets, all indicators show a much more aggressive trend for companies in all industries to transform businesses through software, exploring much further market adjacencies. 54\% of CEOs have entered a new sector or sub-sector, or considered it, in the past three years\footnote{18th Annual Global CEO Survey. PricewaterhouseCoopers. Retrieved from: http://www.pwc.com/gx/en/ceo-survey/2015/download.jhtml}. Also from the same source by PricewaterhouseCoopers, 55\% of entertainment and media CEOs, 52\% of communications CEOs, 48\% of power and utilities CEOs and 47\% of banking and capital markets CEOs say a significant competitor from the technology sector is emerging or will emerge. 

Several factors speed up the growing importance of software. The cloud has become necessary for the survival of technology companies. Revenue for SaaS is expected to grow at a compound annual rate of more than 20\% throughout this decade. By 2018, 59\% of the total cloud workloads will be SaaS\footnote{Global Cloud Index: Forecast and Methodology, 2014–2019. Retrieved from: \url{http://tinyurl.com/q9vxgcx}}. Besides, we are seeing an integration of digital sensors, processing, connectivity and security into virtually every industry's products. As IoT moves out of the hype phase, it will drive demand for network infrastructure, sensors, software applications, and all technologies needed to operate IoT applications including data analytics\footnote{Global technology M\&A 1Q15: first look. Ernst \& Young. Retrieved from: \url{http://www.ey.com/GL/en/Industries/Technology/EY-global-technology-ma-1q15-first-look}; and Manyika, J., Chui, M., Bisson, P., Woetzel, J., Dobbs, R., Bughin, J. and Aharon, D. (2015, June). Unlocking the potential of the Internet of Things. McKinsey Global Institute. Retrieved from: \url{http://tinyurl.com/jqpymqu}}. Cisco predicted that IoT will unleash \$19 trillion USD in new profits and cost savings globally in the next decade (Burrows, 2014). According to Gartner, there will be nearly 26 billion devices on the Internet of Things by 2020, that will potentially generate zetabytes of data annually.

The growth of cloud and IoT poses serious challenges to IT leaders. Developing and deploying smart, connected products and retrofitting existing equipment is very challenging, requiring coordination of network connectivity, application protocols, data analytics, and system management. IoT platforms are being developed\footnote{2015 Technology Industry Outlook. Deloitte. Retrieved from: \url{http://tinyurl.com/o7sb52h}} to simplify the processes of developing, connecting, controlling, and capturing insight from connected products and assets, allowing firms to sense and respond to changing customer needs. In particular, controlling these complex and distributed IoT systems will require advanced Root-Cause Analysis (RCA) capable of precisely synthesizing the status of the system for human beings to make decisions. Specifically, human beings will no longer be capable of controlling so complex system through traditional dashboards and will require a higher level of automation to generate hypotheses of potential root-causes much more accurately.

While several decades of research have produced a large number of algorithms and techniques to perform root cause analysis in many different fields, there is still a lack of understanding on how they can be used and adapted to the growing complexity of IoT and other similar environments, where scalability and real-time reaction become essential. In particular, the appropriate interpretation and management of the vast amounts of data generated in these environments need to be underpinned by IoT and Cloud platforms in order for them to be genuinely viable~\cite{Gubbi20131645}.

In this survey we will thus focus on the RCA models available and the existing generation and inference algorithms that have been developed for them, paying special attention to performance aspects. Although there is a vast literature on RCA, we will restrict to techniques that can be applied to IT systems. Our survey builds on the shoulders of previous contributions from other general surveys like \cite{KavulyaJGN12}. For surveys on specific areas, the reader may refer to \cite{SteinderS04,FengXX13} for computer networks, \cite{Agarwal14,Wong16} for software, \cite{Gao15a,Gao15b} for industrial systems, \cite{Lazarova16} for smart buildings, \cite{Katipamula05} for buildings, \cite{Feng13} for machinery, \cite{Qin14} for swarm systems, \cite{Isermann05,Hwang10} for automatic control systems, \cite{Lanigan11,Mohammadpour11} for automotive systems and \cite{Patton91} for aerospace systems.


The remainder of the paper is organized as follows. Section~\ref{sec:terminology} introduces the main concepts and terminology used in the rest of the paper. Section~\ref{sec:rca_models} describes the main models used for RCA, the different ways in which they can be obtained and some of the learning algorithms available for each model. The inference algorithms that can be used on these models are the subject of Section~\ref{sec:inference} while Section~\ref{sec:conclusion} concludes this paper.

\section{Main concepts and terminology}
\label{sec:terminology}

In this section, we discuss the main concepts and terminology used in the area of RCA. In particular, we provide some essential terminology that we will use throughout the remainder of the survey and provide a classification of RCA tasks.

The core concepts behind RCA are \emph{causality} and \emph{explanation}. Although central to any scientific endeavour, there is no consensus on their formal definition despite centuries long discussions on the subject~\cite{Pearl09,Strevens13}. This is relevant, especially in the case of explanation, because an explanation is often a desired output of RCA, thus, as we will see in Section~\ref{sec:inference}, many alternatives have been proposed.

\subsection{Terminology}
In this survey, we follow the terminology proposed in \cite{KavulyaJGN12}:
\begin{LaTeXdescription}
    \item [Event] is an exceptional condition occurring in the operation of a system. 
    \item [Faults/problems/root causes] are events that can cause other events but are not themselves caused by other events. According to their duration they can be classified as: \emph{permanent} if fault will persist until reparation, \emph{intermittent} if they are discontinuous and periodic, and \emph{transient} if temporary.
    \item [Error] An \emph{Error} is caused by one or more faults and is a discrepancy between a condition of the system and its theoretically correct condition. 
    \item [Failure] A \emph{Failure} is an error that is observable from outside the system. 
    \item [Symptom] A \emph{Symptom} is an external manifestations of failures. This includes a direct observation of failures themselves and externally visible indicators that a failure happened that are not failures by themselves, like alarms raised by anomaly detectors.
    \item [Root Cause Analysis] also referred as \emph{fault localization}, \emph{fault isolation} or \emph{alarm/event correlation}, is the process of inferring the set of faults that generated a given set of symptoms. Note that this process might be trivial if faults are directly observable, in which case they are also symptoms as well. However, this is not the usual case in complex systems. In such cases, a model that explains the relationship between faults and symptoms must be used to be able to perform this inference process.
\end{LaTeXdescription}

\subsection{Classification of the RCA challenges}

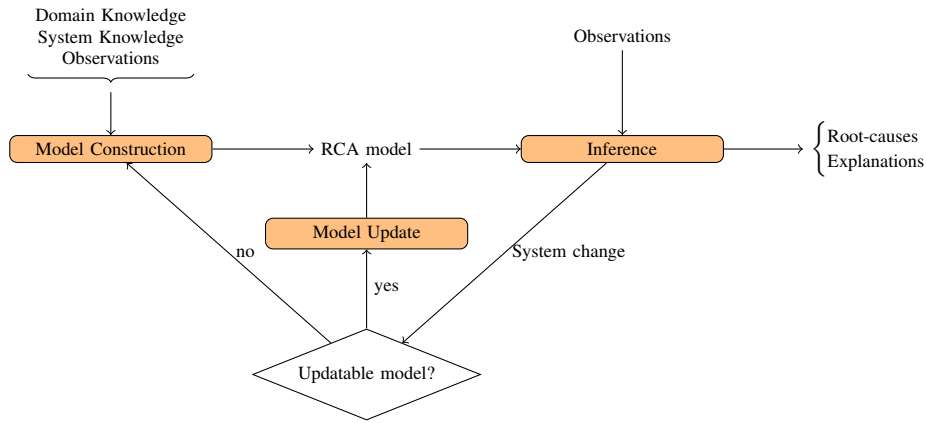
\begin{figure*}[!t]
\centering
\footnotesize
\begin{tikzpicture}[x=1.7cm,y=-1.5cm,every node/.style={font=\scriptsize}]
\tikzstyle{action} = [rounded corners=1mm, text width=2.5cm, draw=black, align=center,fill=orange!50]
\tikzstyle{object} = [draw=none] 
\tikzstyle{choice} = [draw=black, diamond, aspect=2.5] 
\node[action,] (Construction) at (1,1) {Model Construction};
\node[object] (Model) at (3,1) {RCA model};
\node[object,align=center] (Inputs) at (1,0) {Domain Knowledge\\System Knowledge\\Observations};
\node[object] (Observations) at (5,0) {Observations};
\node[action] (Inference) at (5,1) {Inference};
\node[object] (Outputs) at (7,1) {$\begin{cases}\text{Root-causes}\\ \text{Explanations}\end{cases}$};
\node[choice] (Updatable) at (3,3) {Updatable model?};
\node[action] (Update) at (3,1.75) {Model Update};
\draw (Construction) edge[->] (Model);
\draw (Model) edge [->] (Inference);
\draw (Inference) edge [->] (Outputs);
\draw (Inference) edge[->] node[anchor=west] {System change} (Updatable);
\draw (Updatable) edge[->] node[anchor=west] {yes} (Update);
\draw (Update) edge[->] (Model);
\draw (Updatable) edge[->] node[anchor=west]{no} (Construction);
\draw (Observations) edge[->] (Inference);
\draw [decorate,decoration={brace,amplitude=3pt,raise=4pt},yshift=0pt]
([yshift=3pt]Inputs.south east) -- ([yshift=3pt]Inputs.south west) node (Brace) [midway,yshift=-8pt] {};
\draw (Brace) edge[->] (Construction);
\end{tikzpicture}
\normalsize
\caption{Root-Cause Analysis workflow.}
\label{fig:RCA:flow}
\end{figure*}

There is a wide range of RCA models and techniques. One of the reasons that explains such a large corpus is the fact that different aspects and requirements of the system to be analyzed may need different analysis strategies. In this section we describe some of the relevant dimensions that can affect the nature of a RCA problem.
\begin{LaTeXdescription}
\item [Analysis intent] Whether the objective of the analysis is to obtain just the root cause (or causes) of the observed symptoms or an explanation (explaining how the root causes are linked to the symptoms) is desired.
\item [Analysis time] This is a non-functional requirement affecting the maximum time that the algorithm can spend doing inference. Although this requirement can have a numerical translation, it is sometimes useful to use two broad classes: \emph{real-time diagnosis}, in which response time is critical and \emph{post-mortem diagnosis}, in which temporal constraints are not so important. For real-time diagnosis a usual approach is to precompute part or all of the inference process, trading space by online checking time, i.e., the time for analysis still has to be spent during the offline precomputation.
\item [Complexity] Diagnosis can be a more or less challenging task depending on a number of factors:
\begin{LaTeXdescription}
\item [System Size] The number of components in the system to diagnose. This will have an impact in the size of the model used for diagnosis.
\item [Data Size] Volume of data that has to be processed during diagnosis. For diagnosing systems which use data as well to generate the RCA model, this metric can be subdivided between \emph{Training Data Size}, \emph{i.e.},the amount of data that is available for learning the model, and \emph{Observation Data Size}, which is the amount of data that needs to be processed during the inference process. 
\item [Inference length] Maximum number of components that have to be traversed to reach a fault from a symptom. If this length is zero, it means that all faults are symptoms and diagnosis is simply the output of the anomaly detection task. Since the diagnosis of these type of systems is a task of anomaly detection, we will restrict to systems in which the inference length is at least one.
\item [Effect propagation time] Changes in one component of the system may take time to modify the state of adjacent components. In some diagnosing systems the effect propagation time is ignored or considered immediate, like in medical diagnosis where the time window for relevant observations is decided by the doctor, but in other cases the effect propagation time defines the window of relevant observations, thus affecting the Observation Data Size.
\item [Evolution rate] The speed and extent of changes in the diagnosed system. In slow-changing systems (like human health in medical diagnosis applications for a well-established medical area), it might pay off the effort of creating an accurate manual model from experts, since once complete there will be only minor changes to the model.
\end{LaTeXdescription}
\item [Domain knowledge required.] The amount of knowledge on the domain of the system. For instance, if building a diagnosing system for a Hadoop installation, the knowledge on Hadoop that is available to the designer of the diagnosis system.
\item [System knowledge required.] Indicates the level of access the diagnosing system will have to the diagnosed system. This dimension ranges from \emph{black box}, i.e., no information available, to \emph{white box}, i.e., all internal information available, even with the possibility to alter the system. Note that this is orthogonal to the domain knowledge as, for instance, one could have access to the Hadoop code but have no idea on how Hadoop works.
\end{LaTeXdescription}

\subsection{RCA workflow}

Most RCA techniques share a basic workflow that can be summarized in the diagram of Figure~\ref{fig:RCA:flow}. First a model is constructed, combining Domain Knowledge, System Knowledge and  observations of the diagnosed system. Not all these types of informations have to be necessarily used, but approaches exist that can consider all of them in the model construction. 

The output of such process is, obviously, the RCA model that will be used for inference. The model is populated using the observations of the diagnosed system and the outputs, depending on the algorithm and model used can be the root-causes and/or an explanation of the observations.

The type of model has an effect as well on the approach that has to be taken when there is a change in the system (i.e., a change in the System Knowledge) such as the addition/removal of components or connections between them. Some models can be updated incrementally, but for the rest when such changes occur a reconstruction of the model is required. Changes in Domain Knowledge are considered much less frequent than changes in System Knowledge, thus the workflow does not explicitly consider 
Domain Knowledge updates as it often entails a model reconstruction.



\section{Generation of RCA models}
\label{sec:rca_models}
The characteristics of the inference process, in particular its performance, are heavily affected by the type of RCA model used. 
In this section we describe many of the RCA models available in the literature (Section~\ref{sec:rca_models:ssec}) and 
the techniques used to learn/construct these models (Section~\ref{sec:learn_rca_models}).

\subsection{Models for RCA}
\label{sec:rca_models:ssec}
Inference is a fundamental task in Artificial Intelligence (AI) \cite{Russell09}, hence most models for RCA come from this field. 

There are two broad families of models: Deterministic models and Probabilistic models. In deterministic models there is no uncertainty in the known facts or the inferences expressed in the model. On the other hand probabilistic models are able to handle this uncertainty.

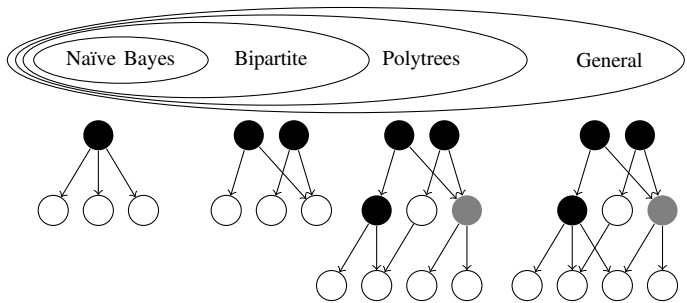
\begin{figure}[!t]
\centering
\footnotesize
\begin{tikzpicture}
\node[ellipse,draw=black] (Naive) at (0,0) {Na\"ive Bayes};
\node (Bipartite) at (2,0) {Bipartite};
\node (Polytrees) at (4,0) {Polytrees};
\node (General) at (6.5,0) {General};
\draw ($(Naive)!0.5!(Bipartite)$) ellipse (2.3cm and 0.5cm);
\draw ($(Naive)!0.5!(Polytrees)$) ellipse (3.4cm and 0.6cm);
\draw ($(Naive)!0.46!(General)$) ellipse (4.5cm and 0.7cm);
\tikzstyle{cause} = [circle,fill=black,minimum size=0.4cm]
\tikzstyle{symptom} = [circle,draw=black,minimum size=0.4cm]
\tikzstyle{intermediate} = [circle,fill=black!50,minimum size=0.4cm]
\begin{scope}[shift={(-0.3,0)}]
\node[cause] (C1) at (0,-1) {};
\node[symptom] (S1) at (-0.6,-2) {};
\node[symptom] (S2) at (0,-2) {};
\node[symptom] (S3) at (0.6,-2) {};
\draw (C1) edge[->] (S1);
\draw (C1) edge[->] (S2);
\draw (C1) edge[->] (S3);
\end{scope}
\begin{scope}[shift={(2,0)}]
\node[cause] (CB1) at (-0.3,-1) {};
\node[cause] (CB2) at (0.3,-1) {};
\node[symptom] (SB1) at (-0.6,-2) {};
\node[symptom] (SB2) at (0,-2) {};
\node[symptom] (SB3) at (0.6,-2) {};
\draw (CB1) edge[->] (SB1);
\draw (CB1) edge[->] (SB3);
\draw (CB2) edge[->] (SB2);
\draw (CB2) edge[->] (SB3);
\end{scope}
\begin{scope}[shift={(4,0)}]
\node[cause] (CP1) at (-0.3,-1) {};
\node[cause] (CP2) at (0.3,-1) {};
\node[cause] (CP1a) at (-0.6,-2) {};
\node[symptom] (SP2) at (0,-2) {};
\node[intermediate] (IP3) at (0.6,-2) {};
\node[symptom] (SP1a) at (-1.2,-3) {};
\node[symptom] (SP1b) at (-0.6,-3) {};
\node[symptom] (SP3) at (0,-3) {};
\node[symptom] (SP3a) at (0.6,-3) {};
\draw (CP1) edge[->] (CP1a);
\draw (CP1a) edge[->] (SP1a);
\draw (CP1a) edge[->] (SP1b);
\draw (CP1) edge[->] (IP3);
\draw (IP3) edge[->] (SP3);
\draw (IP3) edge[->] (SP3a);
\draw (CP2) edge[->] (SP2);
\draw (CP2) edge[->] (IP3);
\draw (SP2) edge[->] (SP1b);
\end{scope}
\begin{scope}[shift={(6.6,0)}]
\node[cause] (CG1) at (-0.3,-1) {};
\node[cause] (CG2) at (0.3,-1) {};
\node[cause] (CG1a) at (-0.6,-2) {};
\node[symptom] (SG2) at (0,-2) {};
\node[intermediate] (IG3) at (0.6,-2) {};
\node[symptom] (SG1a) at (-1.2,-3) {};
\node[symptom] (SG1b) at (-0.6,-3) {};
\node[symptom] (SG3) at (0,-3) {};
\node[symptom] (SG3a) at (0.6,-3) {};
\draw (CG1) edge[->] (CG1a);
\draw (CG1a) edge[->] (SG1a);
\draw (CG1a) edge[->] (SG1b);
\draw (CG1) edge[->] (IG3);
\draw (IG3) edge[->] (SG3);
\draw (IG3) edge[->] (SG3a);
\draw (CG2) edge[->] (SG2);
\draw (CG2) edge[->] (IG3);
\draw (CG1a) edge[->] (SG3);
\draw (SG2) edge[->] (SG1b);
\end{scope}
\end{tikzpicture}
\normalsize
\caption{Hierarchy of Bayesian Network models with an example of each class. Black nodes represent causes, white nodes are symptoms and grey nodes represent nodes that are neither causes or symptoms. These subclasses have been historically important for diagnosis, for instance, a bipartite BN, the QMR-DT \cite{Shwe91} network, was used as expert system for medical diagnosis. Note that, for clarity, not all BN subtypes relevant for diagnosis are represented, for instance BN3M networks \cite{Kraaijeveld05} are general BN with three layers and BN2O \cite{DAmbrosio94} are Bipartite BN in which cause-symptom relations are modeled using a noisy-OR canonical model.}
\label{fig:RCA:BayesianModels}
\end{figure}

These two families comprise several techniques, and each of the techniques can have different implementations with different performance implications. For instance decision trees and neural nets are two different implementations of a classifier and the time for diagnosis of the former is usually faster than the latter. Moreover, inside each model there can be subtypes with specific properties in terms of learning/inference complexity. That is particularly relevant for Bayesian Networks, that can be hierarchically classified as shown in Fig.~\ref{fig:RCA:BayesianModels}. 

\begin{figure*}[!t]
\centering
\footnotesize
\begin{tikzpicture}[x=1.6cm,y=1.5cm,every node/.style={font=\scriptsize}]
\node[text width=1.5cm] (PropositionalLogic) at (1,0) {Propositional Logic};
\node[text width=1.5cm] (FirstOrderLogic) at (0,1) {First Order Logic};
\node[align=center] (AbductiveLogicProgram) at (0,0.1) {Abductive\\Logic\\Program};
\node (FaultTree) at (1,1) {Fault Tree};
\node[text width=1.5cm] (FuzzyLogic) at (0,1.85) {Fuzzy Logic};
\node[align=center] (NonAxiomaticLogic) at (0,2.3) {Non-axiomatic\\Logic};
\node[text width=1.5cm] (PossibilisticLogic) at (1,3) {Possibilistic Logic};
\node[align=center] (DempsterShafer) at (0,3) {Dempster-\\Shafer\\Theory};
\node[align=center] (FuzzyFaultTree) at (1,2) {Fuzzy\\Fault\\Tree};
\node (Codebook) at (4,1) {Codebook};
\node[align=center] (DecisionTree) at (3,0) {Decision\\Tree};
\node (SVM) at (3,1) {SVM};
\node (NeuralNet) at (3,0.5) {Neural Net};
\node[align=center] (BayesianNetwork) at (3,2) {Bayesian\\Network};
\node[text width=1.5cm] (ProbabilisticRelationalModel) at (3,3) {Probabilistic Relational Model};
\node[align=center] (MarkovLogicNetwork) at (2,2.25) {Markov\\Logic\\Network};
\node[align=center] (ArithmeticCircuit) at (4,2) {Arithmetic\\Circuit};
\node[align=center] (BayesianAbductiveLogicProgram) at (2,3) {Bayesian\\Abductive\\Logic\\Program};
\node[text width=1.5cm] (SumProductNetwork) at (4,3) {Sum-Product Network};
\node[text width=1.5cm] (RelationalSumProductNetwork) at (5,3) {Relational Sum-Product Network};
\node[align=center] (HiddenMarkovModel) at (6,2) {Hidden\\Markov\\Model};
\node[align=center] (DynamicBayesianNetwork) at (6,3) {Dynamic\\Bayesian\\Network};
\node (Automaton) at (6.5,0) {Automaton};
\node (PetriNet) at (6.5,1) {Petri Net};
\node[align=center] (StochasticDES) at (7,2) {Stochastic\\DES};
\node[align=center] (StochasticPetriNet) at (7,3) {Stochastic\\Petri Net};
\draw (PossibilisticLogic) edge [<-] (MarkovLogicNetwork);
\draw (PropositionalLogic) edge[<-] (BayesianNetwork);
\draw (FirstOrderLogic) edge[->] (BayesianNetwork);
\draw (FuzzyFaultTree) edge[->, bend right=10] (BayesianNetwork);
\draw (FaultTree) edge[->] (BayesianNetwork);
\draw (PropositionalLogic) edge[<-] (DecisionTree);
\draw (PropositionalLogic) edge[<-] (SVM);
\draw (PropositionalLogic) edge[<-] (NeuralNet);
\draw (FuzzyLogic) edge[<-] (NeuralNet);
\draw (ArithmeticCircuit) edge[<->] (SumProductNetwork);
\draw (ArithmeticCircuit) edge[<-] (BayesianNetwork);
\draw (Automaton) edge[<->] (PetriNet);
\draw (ProbabilisticRelationalModel) edge [->] (ArithmeticCircuit);
\begin{scope}[on background layer]
\node[draw,label={[rotate=90,anchor=south]left:{\emph{Deterministic}}},fit=(PropositionalLogic) (FirstOrderLogic) (FaultTree) (Codebook) (DecisionTree) (SVM) (NeuralNet) (Automaton) (PetriNet), inner xsep=10pt, inner ysep=0pt] (Deterministic) {}; 
\node[draw,label=below:{\emph{Logic}}, fill=black!50, opacity=0.2, rounded corners=0.2cm, fit=(PropositionalLogic) (FirstOrderLogic) (FaultTree) (FuzzyLogic) (DempsterShafer) (FuzzyFaultTree) (MarkovLogicNetwork)] (Logic) {};
\node[draw,label={[rotate=90,anchor=south]left:{\emph{Probabilistic}}},fit=(FuzzyLogic) (DempsterShafer) (FuzzyFaultTree) (BayesianNetwork) (ProbabilisticRelationalModel) (MarkovLogicNetwork) (ArithmeticCircuit) (BayesianAbductiveLogicProgram) (SumProductNetwork) (RelationalSumProductNetwork) (HiddenMarkovModel) (DynamicBayesianNetwork) (StochasticDES) (StochasticPetriNet), inner ysep=15pt, inner xsep=10pt] (Probabilistic) {}; 
\node[draw,label=below:{\emph{Compiled}}, fill=black!50, opacity=0.2, rounded corners=0.2cm, fit=(Codebook) (ArithmeticCircuit)] (Compiled) {}; 
\node[draw,fill=black!50, opacity=0.2, rounded corners=0.2cm,label=below:{\emph{Classifier}}, fit=(DecisionTree) (SVM) (NeuralNet) (BayesianNetwork), inner xsep=-2pt] (Classifier) {}; 
\node[draw,fill=black!50, opacity=0.2, label=above:{\emph{Bayesian}}, rounded corners=0.2cm, fit=(BayesianNetwork) (ProbabilisticRelationalModel) (MarkovLogicNetwork) (ArithmeticCircuit) (BayesianAbductiveLogicProgram) (SumProductNetwork) (RelationalSumProductNetwork) (HiddenMarkovModel) (DynamicBayesianNetwork)] (Bayesian) {};
\node[draw,fill=black!50, opacity=0.2,label=below:{\emph{Process Model}},rounded corners=0.2cm, fit=(HiddenMarkovModel) (DynamicBayesianNetwork) (Automaton) (PetriNet) (StochasticDES) (StochasticPetriNet)] (ProcessModel) {};
\end{scope}
\end{tikzpicture}
\normalsize
\caption{Classification of RCA models. Directed edges indicate possible conversions between models.}
\label{fig:RCA:models}
\end{figure*}
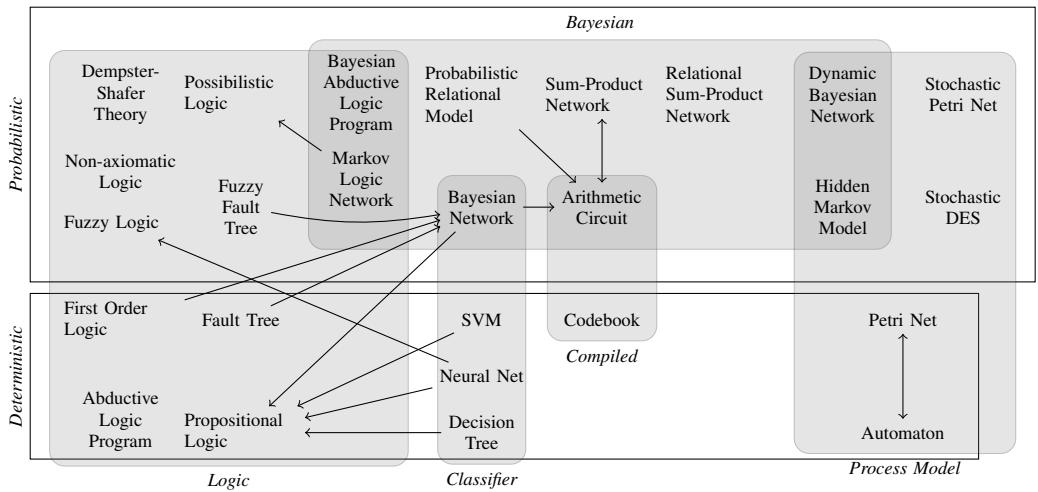

\begin{table}
\caption{Models for RCA}\label{table:RCA:models}
{
\centering
\scriptsize
\begin{tabular}{l|l|l|l}
Family & Technique & Implementation & Used for\\
        &&&diagnosis\\
\hline
Deterministic   & Logic         & Propositional Logic (rule sets) & \cite{Buchanan84,Buning99}\\
                &               & First-order Logic & \cite{Smith89,Daily2012}\\
                &               & Fault Tree & \cite{Duan12,Suntoyo15}\\
                &               & Abductive Logic Programs & \cite{Ciampolini04,Perri05,Zawawy12}\\
                \cline{2-4}
                & Compiled    & Codebooks & \cite{Yemini96,Reali09}\\
                \cline{2-4}
                & Classifier    & Decision Tree & \cite{Zheng04}\\
                &               & SVM & \cite{Demetgul13,Ye14}\\
                &               & Neural Net & \cite{Sorsa93,Alaeddini11}\\
                \cline{2-4}
                & Process  & Automata/FSM & \cite{Sampath96}\\
                & Model              & Petri Nets & \cite{Genc2003,Adamyan04}\\
\hline
Probabilistic   & Logic         & Fuzzy Logic & \cite{Frank97,Sanchez}\\
                &               & Dempster–-Shafer theory & \cite{Yang06}\\
                &               & Fuzzy Fault Tree & \cite{Peng08,Senol15}\\
                &               & Possibilistic Logic \cite{Dubois94} & \cite{Cayrac95}\\
                &               & Non-axiomatic Logic & \cite{Wang07}\\
                \cline{2-4}
                & Bayesian      & Bayesian Networks & \\
                &               & \hspace{2em}Na\"ive Bayes & \cite{Yan10}\\
                &               & \hspace{2em}Bipartite & \cite{Shwe91}\\
                &               & \hspace{2em}Polytree & \hspace{1em}?\\
                &               & \hspace{2em}General & \cite{Kirsch94,Alaeddini11}\\
                &               & Probabilistic Relational Models & \cite{Lee06}\\
                &               & Bayesian Abductive Logic Programs& \cite{Suntoyo15}\\
                &               & Markov Logic Networks & \cite{Zawawy12,Schoenfisch15}\\
                &               & Sum-Product Networks & \hspace{1em}?\\ 
                &               & Relational Sum-Product Networks & \cite{NathD16}\\
                &               & Dynamic Bayesian Networks & \cite{Jha09,Yu13}\\
                &               & Hidden Markov Models & \cite{Salfner07,LiSha13}\\
                \cline{2-4}
                & Compiled      & Arithmetic Circuits & \cite{Duan13} \\
                \cline{2-4}
                & Classifier    & Bayesian MSVM \cite{Zhang06}, LS-WSVM& \cite{Ye08}\\
                &               & Probabilistic Neural Net & \cite{Xu10,Er12}\\
                \cline{2-4}
                & Process  & Stochastic DES & \cite{Dutta15}\\
                & Model              & Stochastic Petri Nets & \cite{Aghasaryan98}\\
\end{tabular}}
\normalsize
\end{table}

Table~\ref{table:RCA:models} shows some of the most well-known models for RCA, together with an example reference in which each model was used for diagnosis. For some models we were not able to find any diagnostic application in the literature, a fact that we have denoted by a "?" in the table. Note, however, that one of these models are Sum-Product networks, which are closely related to Arithmetic Circuits, thus \cite{Duan13} would be a valid reference for them. 

The table is not complete in the sense that any classifier can be used for RCA (at least for single-fault diagnosis) as long as there is enough training data. Training data would be instances of symptoms with their corresponding fault(s) as label. Since the number of different classifiers currently available is quite large, we only mention a subset of them in this table. Besides, some of them are mentioned both in the deterministic and the probabilistic families since, depending on the codification of the problem and their construction, they might be able to handle uncertainty on inputs and provide confidence values on outputs.

Although classifiers are attractive specially because their automatic generation has been one of the key researched topics in Machine Learning, they do not dominate the RCA area. Some of the reasons that can help explain this effect are that the majority of the most advanced classifiers, like Neural Nets, (i) only return a predicted root cause and it is difficult to obtain an explanation from them. (ii) Do not yield  logical rules, and such approaches are difficult to combine with available domain knowledge, although not impossible \cite{Sima95}. (iii) They are usually tailored for single label classification, which would correspond to a single fault diagnosis task. If multiple-fault diagnosis has to be achieved, then a selection strategy has to be implemented to generate the set of faults out of a multi-class classifier (e.g., like taking all labels above a defined threshold) or work with multi-label classifiers (see \cite{Zhang14} for a good survey on the area).

Using a table to create a taxonomy of RCA models is helpful to mentally order the landscape of available models, but hides the fact that relationships between models are not as clean as they might seem. For instance, codebooks~\cite{Reali09,Yemini96} can be seen as a particular implementation of propositional logic, as they are basically a way to precompute the inference on top of a graph by generating sets of rules that can be quickly checked using a mechanism such as hash tables. Similarly, we have established a distinction between models able to diagnose situations in which time of observation of symptoms is not relevant for inference, and process models which explicitly consider the sequence of observations. However, there are process models that are Bayesian approaches as well, like Dynamic Bayesian Networks or Hidden Markov Models. These relationships can be better appreciated in the diagram of Figure~\ref{fig:RCA:models}.

If domain knowledge is provided in a given model, but the preferred model for inference is different, there are ways in which one model can be (sometimes losslessly) converted into another. For instance sets of (in some cases fuzzy) rules can be extracted from decision trees \cite{Quinlan93}, Bayesian Networks \cite{Hruschka07}, SVMs \cite{Barakat10} and Neural Nets \cite{Andrews95,Chen05}, Possibilistic Logic derived from Markov Logic Networks \cite{Kuzelka15}, Bayesian networks can be generated from first-order logic \cite{Haddawy94} or fault trees \cite{Bobbio01} if probabilities are provided, fuzzy fault trees \cite{Wang12} and Sum-Product Networks \cite{Zhao15}, and Arithmetic circuits and Sum-Product Networks can be converted one into the other~\cite{rooshenas14}. Some of these conversions have a strong effect on diagnosis performance, for instance Bayesian Networks and Relational Bayesian Network (one type of Probabilistic Relational Model) can be compiled into Arithmetic Circuits for particular diagnosing tasks \cite{Chavira06}. The compilation process is expensive, but after that the diagnosis process is much faster \cite{Darwiche09}.


Models have their own characteristics, which can have a large impact on the diagnosis performance: 
\begin{LaTeXdescription}
\item [Size.] The number of elemental analysis elements used to model the system (typically number of components, but the exact definition depends on the abstraction level at which the system is modeled). Depending on the model this can be the number of variables,  rules, nodes, etc. Models might have more than one size attribute.
\item [Inference structure.] Defines how the different elemental analysis elements (rules, nodes in a Bayesian Network, etc.) relate to each other. Many techniques are tailored for specific structures, as structure can have a large impact on the theoretical complexity as well as final performance. Since it is complex to succinctly specify the structure of a model, a derived metric, \textbf{Inference length}, is useful to distinguish between diagnosing techniques. Inference length is the maximum number of inference steps needed to reach a fault from a symptom in the model. 
\end{LaTeXdescription}

There are three main ways in which a model for RCA can be obtained:
\begin{LaTeXdescription}
\item[Manual generation] In this case a group of experts provides the model. Models produced in this way tend to be very accurate, but knowledge elicitation is a complex and slow process \cite{Korb10}. For systems with a high evolution rate this approach might be unpractical.
\item[Assisted generation] In most of the cases domain knowledge is partially available, for instance in the form of known models for sub-parts of the system that can be replicated several times and arbitrarily connected to other sub-parts. In these cases, the whole model is produced by assembling the models of the sub-parts based on available data of the system (i.e., available system knowledge), like its topology. In some cases these sub-models are not explicit in a library but implicit in the algorithm that, given the available system knowledge, generates the model for RCA. Assisted generation methods require a fair amount of domain knowledge in the form of a submodel library and/or the specific composition algorithms, plus a detailed system information to be able to produce the final diagnosis model. Most of the RCA systems in the literature applied to industrial environments fall inside this category, as it offers a good compromise between the quality of manual specification and the automatization of the construction of the whole model.
\item[Automated generation] The model for RCA is generated entirely from the data, using standard non adhoc algorithms, which may include observations as well as available system knowledge. This is the only viable solution if it is not possible or practical to obtain domain knowledge.
\end{LaTeXdescription}


\subsection{Learning models for RCA}
\label{sec:learn_rca_models}
When no domain knowledge is available to generate a model, either manually or in some assisted way, the only remaining options is to use learning algorithms on the raw data of the system. For models that are classifiers, this is quite straightforward since they originated in the Machine Learning area. For models that had different origins, like Bayesian Networks, a wide range of techniques have been developed to learn the models from the data. Table~\ref{table:RCA:learning} summarizes the algorithms available for that task. 
\begin{table}
\caption{Automated construction algorithms of RCA models}
\label{table:RCA:learning}
\begin{center}
{
\centering
\begin{tabular}{l|l}
Model & Learning Algorithm \\
\hline
Fault Tree & \cite{Liggesmeyer98,Majdara09,Hussain10,Venceslau14,Roth15}\\
\hline
Dempster-Shafer Theory & --\\
Fuzzy Fault Tree & --\\
First Order Logic & Inductive Logic Programming \cite{Muggleton94} \\
    & \cite{Augier95,Tamaddoni01}\\
\hline
Markov Logic Network & L-BFGS \cite{Richardson06}\\
                     & LHL \cite{Kok09}\\
\hline
Hidden Markov Model & RJMCMC \cite{Robert00} \\
\hline
Dynamic Bayesian Network & \cite{Lahdesmaki08}\\
                         & MCMC sampling \cite{Robinson10}\\
\hline
Probabilistic Relational Model & \cite{Friedman99}\\
Stochastic Petri Net & \cite{RoggeSolti13}\\
\hline
Codebooks & \cite{Yemini96} \\
\hline
Decision Trees  & ID3 \cite{Quinlan86} \\
                & CHAID \cite{Kass80}\\
                & C4.5 \cite{Quinlan93}\\
                & ID5R \cite{Utgoff89}\\
                & MinEntropy \cite{Zheng04}\\
                & Conditional Inference trees \cite{Hothorn06}\\
\hline
Propositional logic (rule sets) & Separate-and-conquer \cite{Furnkranz99}\\
                & Association rules \cite{Agrawal93}\\
               & Artificial Immune Systems \cite{Aydin10}\\
\hline
Fuzzy logic (fuzzy rule sets)& Artificial Immune Systems \cite{Mezyk11}\\
Possibilistic Logic & \cite{Serrurier07,Kuzelka16}\\
Non-axiomatic Logic & --\\
\hline
Automata/FSM & See \cite{Cook98} \\
\hline
Petri nets & See \cite{AalstD13}\\
\hline
Na\"ive Bayes & NBE \cite{Lowd05}\\
\hline
Bipartite BN & Iterative SEQ \cite{Kearns98}\\
\hline
Polytree BN & ?\\
\hline
(General) Bayesian Networks & IC, IC* \cite{Pearl09}\\
                   & PC \cite{Spirtes00}\\
                   & Dynamic Programming \cite{silander06}\\
                   & POPS \cite{Fan14}\\
                   & parallel PC \cite{LeHL0L15,MadsenJSLN15}\\
                   & MDL \cite{Suzuki93}\\
                   & B\&B MDL \cite{Tian00}\\
                   & K2 \cite{Cooper92}\\
                   & FCI \cite{Spirtes00}\\
                   & MMHC \cite{Tsamardinos06}\\
                   & GLL, LGL \cite{Aliferis10}\\ 
                   & $\chi^2$ test on updates \cite{Bennacer15}\\
\hline
Arithmetic circuits  & \cite{Lowd12}\\ 
\hline
Sum-product networks & \cite{GensD13}\\
\hline
Relational Sum-product networks & \cite{NathD15}\\
    \hline
     Neural Nets & See \cite{Rojas96}\\
    \hline
    SVM & See \cite{bottou07}\\
\end{tabular}}
\end{center}
\end{table}

Some models are complex enough to distinguish between their structure and the values of their internal parameters. This is the case for instance for BNs or Fuzzy Logic in which the structure corresponds to how variables are related (through arcs in BNs and through clauses in Fuzzy Logic) and the probabilities assigned to the elements in that structure are the parameters. Learning algorithms can learn both things (structure and parameters) or just one of them. In general in the table we have listed algorithms that learn everything or just the structure, as they frequently are more complex than the ones just learning the parameters given the structure. Note, however, that for Neural nets a structure is usually assumed and learning algorithms in that case usually refers to parameter learning algorithms. 

In some cases no specific learning algorithm was found in the literature (e.g., Dempster-Shafer Theory, Fuzzy Fault Tree and Non-axiomatic Logic), although this does not preclude the option of learning an alternative model and then using a conversion between models as the one reported in Section~\ref{sec:rca_models:ssec}.

In some other cases the referenced algorithms learn specific subclasses of the model. For instance, to our knowledge, there are no general algorithms for learning Bipartite BNs. The only related approach is \cite{Kearns98}, where the Bipartite BN had to satisfy additional constraints, like using a noisy-Or model \cite{Pearl86}, that each symptom could be only related to a maximum of $k$ different causes, and that the probabilities could not be arbitrary. In that case, their algorithm learns the BN with cost exponential on $k$, but linear in the number of causes multiplied by the number of symptoms. Similarly, we are not aware of any specific learning algorithm for the class of polytree BNs, other than using a learning algorithm for general BNs trying to  constrain the decisions in the learning process to comply with the acyclic property of polytree BNs. 



Among the fastest learning algorithms are the ones learning rules (either decision trees or directly rule sets) for non-process models. Some of them are even greedy in order to speed up diagnosis in production systems, like \emph{MinEntropy} \cite{Zheng04}, which was used at eBay. For process models many fast algorithms exist to obtain automata~\cite{AalstRVDKG10} or Petri nets~\cite{AalstWM04}, although not very accurate in many cases.

Some interesting work focuses in creating self-adaptive decision support trees based on streamed data including change detection. Interestingly, this work assumes that the amount of information is so large that the data stream can be considered continuous and in general it is not possible to read data more than once. These assumptions are specially well suited when dealing with IoT systems or other complex systems that change dynamically, which cannot be fully understood in general because of their size and the size of the data generated. Differently from classical methods such as C4.5, these algorithms do not assume that all training data are available simultaneously in memory and deal with change over time. In particular, different methods have been proposed based on Hoeffding Trees or Very Fast Decision Tree method (VFDT)~\cite{Domingos00}. For instance, CVFDT is an adaptive variant of VFDT proposed by Hulten et al~\cite{Hulten01}. Adaptive Hoeffding Trees were later proposed in~\cite{Bifet2009} for the same purpose, but detecting change and updating the decision tree based exclusively on data analysis. Attempts to parallelize the creation of a decision tree for heavy streams are implemented in Apache Samoa\footnote{Apache Samoa: https://samoa.incubator.apache.org}.   

Other classifier-based RCA models in the table, like Na\"ive Bayes, SVMs or Neural Nets have very different training costs. While Na\"ive Bayes can be trained in a single pass of the data, thus, according to \cite{Domingos00}, it was one of the most widely used learners at Google, the
complexity of general (non-linear) SVM classifiers during training is between $O(n^2)$ and $O(n^3)$, where $n$ is the number of training instances, although with an iterative approximate approach \cite{Tsang05} it can go down to $O(nr)$, where $r$ is the number of iterations performed. On the other hand, the general learning problem for Neural nets is NP-complete \cite{Blum92}. 

In this regard, one of the most popular RCA models, BNs do not enjoy any advantage with respect to classifier-based models: 
learning Bayesian Networks is NP-complete \cite{Chickering96}\footnote{Learning algorithms for BNs can be classified into constraint-based learners and score-based ones. Although this is sometimes cited as a general rule, in fact \cite{Chickering96} shows that the methods based on scores and search are NP-complete}. However there has been an evolution in the sizes of the systems whose causality can be inferred. For instance, the complexity of the PC algorithm, one of the first algorithms to be used, is $O(n \log(n) \max{(p^q, p^2)})$, where $n$ is the number of samples, $p$ is the number of variables, and $q$ is the maximal size of the adjacency sets \cite{KalBue08}, thus worst-case exponential. This made it quite difficult to use it for more than a hundred of variables. More recent approaches, like LGL, based on generating global causality from local causality have raised this limit to one million variables, though running times can range from quadratic time to exponential time depending on parameters provided to the algorithm~\cite{Aliferis10b}.

Besides these improvements based on changing the philosophy of the learning algorithm, there have been proposals as well in the line of parallelizing existing algorithms, like \emph{parallel PC} \cite{LeHL0L15,MadsenJSLN15}. Although we only mention this parallel algorithm in the table, the possibility of parallelization has also been explicitly considered for many of the other algorithms\cite{CamposJ11,Aliferis10b}\footnote{In principle any algorithm can be parallelized, albeit with different success in terms of being close to the theoretical maximum speedup.}. 

If the evolution rate of the diagnosed system is high, then this will impose restrictions on the algorithms and/or the models for RCA that can be used. Either the model/learning algorithm allows for incremental changes or it is fast enough to learn the whole model from scratch every time there is a change. For instance algorithms \emph{ID5R} for decision trees and \emph{$\chi^2$ test on updates} for Bayesian Networks can work incrementally.

Classifiers in general are not a good option when the evolution rate of the diagnosed system is high, specially if the classifier requires lots of instances to attain good results. There are several factors that explain this fact: First, if evolution rate is high, then it is likely that there will be few instances available to train the classifier. Second, assuming that the system is large enough so that fast changes produce a reasonable quantity of diagnosable instances, to correctly label them we will need human diagnosis. Typically this is a slow process (the worth of the automated RCA system is in fact directly related to how time consuming is the manual diagnosis), so throughput could only be increased by having a large pool of humans. Crowdsourcing has been proposed as a solution when large amounts of workers are required, however in this case the task is highly specialized and most of the times involves dealing with sensitive data, thus not very suitable for this approach. A possible way to compensate for scarcity of experts could be systems providing guesses that have to be later on validated by humans. However this approach relies on the assumption that validation is significantly faster than diagnosis, which depends on the feedback provided by the classifier. If just a guessed label is given and nothing else, then the savings would not be as large as if additional information, like the followed reasoning or the facts that support that hypothesis are given to the user.


\section{Inference in RCA models}
\label{sec:inference}

Once the model for RCA is available, it is possible to use it to obtain the fault (or faults) that generate a given set of symptoms. This process is called \emph{inference} or \emph{abduction}. For classifiers, the process is straightforward as it simply involves the classification of the symptoms using the model. For other models, Tables~\ref{table:inference:nonbayesian} and \ref{table:inference:bayesian} show several algorithms available to find the root causes given the set of symptoms. 

In this table there are two types of algorithms: algorithms that provide equivalent results but with different implementations offering different performances (that is the case of the Rete family of algorithms), and algorithms that have a different concept of what provides a good explanation.

For rule sets, the concept of a good explanation is relatively simple: they can provide the conclusion to the user, together with all the rules fired to reach that conclusion from the symptoms. On the other hand, for Bayesian Networks the concept is not so clear. One possibility is to use the computed probabilities of each of the potential causes, their \emph{marginals}, and take the one with maximum probability (or the group above a given threshold if multiple faults are allowed). However this approach considers the aggregation of all possible worlds compatible with the observations, even if the real world can only be one of those states. In contrast, the \emph{Most Probable Explanation} (MPE) just outputs the most probable compatible world, that is the assignment of all the variables in the model with highest probability, thus it is most meaningful when the model contains only causes and symptoms and all symptom values are known. The \emph{Maximum A Posteriori} (MAP) lies somewhere in between the previous two: some of the variables can be abstracted, thus aggregating some of the worlds, before selecting the most probable one. This is done typically with intermediate variables not representing the potential causes or with symptoms for which we have no information, so that only the set of all potential causes is considered. 

Recently new metrics have been proposed, like the \emph{Most Reasonable Explanation} (MRE) or the \emph{Most Inforbable Explanation} (MIE). All of them try to find, according to some goodness criteria, what subset of variables is the most informative to the user to explain the observations, rather than the user having to provide that set beforehand as in MAP. 
However this more sophisticated explanations are not needed if the model has clear variables and/or values that are tied to failures, and the number of this potential causes is not very large. For instance, if the model shows the dependency between components and each component has an associated variable with two possible values, \emph{ok} and \emph{failure}, given some symptoms, computing the posterior marginal probabilities for each one of the variables in its failure state, or computing the MPE or the MAP would be a good option. The particular selection between Marginals, MEP or MAP depends on whether variables not related to failures have to be abstracted or not and the number of faults that can happen concurrently: marginals would be a better option for single-fault diagnosis, while MPE and MAP are more suited for multiple failure scenarios. On the other hand, if variables represent simply different possible configuration options and none of them is inherently wrong, alternative explanation methods will try to automatically subset which variables had the largest influence in the observed symptoms.

In Tables~\ref{table:inference:nonbayesian} and~\ref{table:inference:bayesian} we have selected the following dimensions to classify the algorithms:
\begin{LaTeXdescription}
\item[Multiple Failure] If algorithm is capable of finding several concurrent failures (\cmark) or just returns always at most a single failure (\xmark). Some methods only work under the assumption that there can be at most $k$ different concurrent failures. This is indicated in the table with a $k$.
\item[Exact] If the algorithm yields the exact answer to the corresponding inference concept (\cmark) or not (\xmark). A negative answer could be either because the algorithm uses some heuristic to reduce the search space or because it is an iterative method that converges to the correct answer with time, suitable if an anytime algorithm is required.
\item[Unknown Symptoms] The method is able to diagnose with incomplete symptom information (\cmark). On the other hand, (\xmark) indicates that the algorithm requires the values for the symptom variables to be all available, i.e., all symptoms are either positive or negative.
\item[Unknown Causes] The algorithm doesn't need to know which variables (or variables values) correspond to potential root causes/failures, so it has to evaluate the contribution of each variable in the outcome to decide which subset of variables had a greater impact in generating the symptoms (\cmark). If potential failure variables need to be known and have identified failure values, this is denoted by (\xmark).
\item[Noisy Symptoms] Positive/negative symptom information can be wrong and the algorithm can still provide a reasonable (depending on the amount of noise) answer (\cmark). If algorithm needs precise symptom information, then (\xmark).
\item[Noisy Propagation] The algorithm is able to consider that failures do not propagate perfectly: there is a chance that errors can be masked in a part of the system and do not propagate to other connected elements (\cmark).
\item[Adaptive] Solving an instance of the problem provides an internal state and a solution that can be reused for similar queries or as new information (evidence) is available.
\end{LaTeXdescription}
\begin{table*}
\caption{Inference algorithms in non-Bayesian RCA models. $r$ is the number of rules, $e$ are the number of evidences (symptoms or any other type), $c$ is the average number of conditions per rule, $f$ the number of potential faults, $a$ arithmetic circuit/SPN size, $d$ nodes of decision tree, $l$ is the number of layers of the neural net, $h$ is the number of neurons in a hidden layer.} \label{table:inference:nonbayesian}
\begin{center}
{
\scriptsize
\begin{tabular}{l|l|l|c|c|@{\hskip3pt}c@{\hskip3pt}|c|c|c|c|l}
&&&&&&&&&\\
&&&&&&&&&\\
&&&&&&&&&\\
&&&&&&&&&\\
Model & Inference Concept & Inference Algorithm  
&\begin{rotate}{90}Multiple Failure\end{rotate}
&\begin{rotate}{90}Exact\end{rotate} 
&\rotatebox{90}{\shortstack[l]{Unknown\\Symptoms}}
&\begin{rotate}{90}Unknown Causes\end{rotate} 
&\begin{rotate}{90}Noisy symptoms\end{rotate} &\begin{rotate}{90}Noisy propagation\end{rotate}
&\begin{rotate}{90}Adaptive\end{rotate}
& Cost\\
\hline
Codebooks & Symptom Vector  & Min Hamming Distance Decoder \cite{Yemini96} & \xmark & \cmark & \xmark & \xmark & \cmark & \xmark & \xmark & $O(f\cdot s)$\\
         & Similarity & Multiple fault detection \cite{Monacelli11} & $k$ & \cmark & \xmark & \xmark & \cmark & \xmark & \xmark & $O(f^k\cdot s)$\\
\hline
    Propositional logic  & Logic Abduction 
    & Resolution & \cmark & \cmark & \cmark & \xmark & \xmark & \xmark & \xmark & $O(\exp(r+e))$\\
\hline
    First-order Logic  & Forward chaining & Rete \cite{Forgy82} & \cmark & \cmark & \cmark & \xmark & \xmark & \xmark & ? & $O(r e^c)$\\    
            (rule sets without recursive functions)& & Rete-II  &\cmark & \cmark & \cmark & \xmark & \xmark & \xmark & ? &? $(\leq O(r e^c))$\\
            && Rete-III&\cmark & \cmark & \cmark & \xmark & \xmark & \xmark & ? &? $(\leq O(r e^c))$\\
            && Rete-NT &\cmark & \cmark & \cmark & \xmark & \xmark & \xmark & ? &? $(\leq O(r e^c))$\\
            & & Collection Oriented Match \cite{Acharya93}&\cmark & \cmark & \cmark & \xmark & \xmark & \xmark & ? &? $(\leq O(r e^c))$\\
            && RETE* \cite{Wright03}&\cmark & \cmark & \cmark & \xmark & \xmark & \xmark & ? &? $(\leq O(r e^c))$\\
\hline
    First-order Logic & Logic Abduction & Tableau, Sequent calculi \cite{Mayer93}&\cmark &\cmark&\xmark&\xmark&\xmark&\xmark&\xmark  & undecidable \\
\hline
Possibilistic Logic & Logic Abduction & \cite{Cayrac95} & \cmark & \cmark & \cmark & \cmark & \cmark & \cmark & \xmark &? \\
\hline
Dempster-Shafer Theory & MPE & \cite{shafer76} & \cmark & \cmark & \cmark & \cmark & \cmark & \cmark & \xmark & 
$O(2^{\min(m,\|\Theta\|)})$\\
&  & Monte-Carlo \cite{Wilson91} & \cmark & \xmark & \cmark & \cmark & \cmark & \cmark & \xmark & $O(m\|\Theta\|)$\\
\hline
Fault Tree & Tree search & Fault tree top-down \cite{Xu14} & \cmark & \cmark & \xmark & \xmark & \xmark & \xmark & \xmark & $O(e+f)$ \\
\hline
Decision trees       
            & Explanation  & Four heuristics \cite{Zheng04}  & \cmark& \xmark & \xmark & \cmark & \xmark & \xmark & \xmark & $O(d^2)?$\\
            & Simplification                         & Decision to largest number of faults \cite{Chen04}& \cmark & \cmark & \xmark & \cmark & \xmark & \xmark &\xmark &? ($\leq O(d^2)$)\\
            \cline{2-11}
           & Most likely decision & Decision tree easiest mutation \cite{Aggarwal09} & \xmark& \cmark & \xmark & \cmark & \xmark & \xmark & \xmark & $O(d^2)?$\\ 
\hline
Sum-product networks & Marginals & \cite{Poon11} & \xmark & \cmark & \cmark & \xmark & \cmark & \cmark & \xmark & $O(a)$\\
& MPE & \cite{Poon11} & \cmark & \cmark & \cmark & \xmark & \cmark & \cmark & \xmark & $O(a)$\\
\hline
Neural nets &Classification&Single class classifier& \xmark& \cmark & \xmark & \xmark & \cmark & \xmark & \xmark & $O(\max(s,f,h)^2\cdot l)$\\
 & &Multiple class classifier& \cmark& \cmark & \xmark & \xmark & \cmark & \xmark & \xmark & $O(\max(s,f,h)^2\cdot l)$\\
\hline
SVM &Classification&Single class classifier& \xmark& \cmark & \xmark & \xmark & \cmark & \xmark & \xmark & kernel dependent\\
\hline                       
Markov Logic Networks   
    & Logic abduction & Pairwise Constraints \cite{kate09,singla11} & \cmark & \cmark & \cmark & \xmark & \cmark &\cmark & \xmark & $O(\exp(w))$ \\ 
    \cline{2-11}
                        & Marginals & FOVE \cite{Braz07} &  \xmark & \cmark & \cmark & \xmark & \cmark &\cmark & \xmark & ?\\ 
                        & & WFOMC \cite{Broeck11} &\xmark & \cmark & \cmark & \xmark & \cmark &\cmark & \xmark & $O(\text{formula size})$ \\
                        && Compilation to C++\cite{Kazemi16} &\xmark & \cmark & \cmark & \xmark & \cmark &\cmark & \xmark & $O(\text{formula size})$\\
    \hline
    Petri Nets & Most likely sequence& Viterbi puzzle \cite{Aghasaryan98} & \cmark& \cmark & \cmark & \xmark & \cmark & \cmark & \xmark & ?\\
\hline
    SRLG, & Set covering  & Minimum set covering \cite{Kiciman05} & \cmark & \cmark& \xmark & \xmark & \xmark & \xmark& \xmark& NP-hard\\  
    Symptom fault map   &    (non probabilistic)     & Greedy set covering \cite{Kompella05} & \cmark & \xmark & \xmark & \xmark & \cmark & \xmark &\xmark& $O(e\cdot f)$\\
                        
\hline
Arithmetic Circuits    &MPE & MPE (AC) \cite{Darwiche09} & \cmark & \cmark & \cmark & \xmark & \cmark & \cmark & \xmark & $O(a)$ \\
                       &MAP & MAP (AC) \cite{HuangCD06} & \cmark & \cmark & \cmark & \xmark & \cmark & \cmark & \xmark & ?

\end{tabular}\normalsize}
\end{center}
\end{table*}

In the tables some of these characteristics have a more subtle interpretation than the binary one that appears in it suggests. For instance, in rule-based approaches, the approximation is exact as long as there are no conflicts (in which case the output will depend on the conflict resolution algorithm used) and typically one would assume that symptoms have to be known, as in propositional/first-order logic facts have to be either true or false. However, these systems were used for expert system design for medical diagnosis (the famous MYCIN system) and had to support lack of information. Typically this can be achieved by introducing variables to represent if the value of a specific variable is known or unknown and taking into account specifically this possibility in the rules, albeit at the cost of having more rules in place (see \cite{Onisko01} for a comparative survey between rule-based and Bayesian approaches in the context of medical diagnosis). 

\begin{table*}
\caption{Inference algorithms in Bayesian Networks. $n$ is the number of variables, $m$ is the number of variables that change, $w$ is the treewidth of the network, $w_c$ is the constrained treewidth, $d$ average domain size of a variable, $s$ is the number of sympotm variables and $f$ is the number of potential fault variables.}
\label{table:inference:bayesian}
\begin{center}
{
\scriptsize
\begin{tabular}{l|l|l|c|c|@{\hskip3pt}c@{\hskip3pt}|c|c|c|c|l}
&&&&&&&&&\\
&&&&&&&&&\\
&&&&&&&&&\\
Model &  \shortstack{Inference\\Concept} & Inference Algorithm  &\begin{rotate}{90}Multiple Failure\end{rotate}
&\begin{rotate}{90}Exact\end{rotate} 
&\rotatebox{90}{\shortstack[l]{Unknown\\Symptoms}}
&\begin{rotate}{90}Unknown Causes\end{rotate} 
&\begin{rotate}{90}Noisy symptoms\end{rotate} &\begin{rotate}{90}Noisy propagation\end{rotate}
&\begin{rotate}{90}Adaptive\end{rotate}
& Cost\\
\hline
Na\"ive Bayes & Marginals & Marginals & \xmark & \cmark & \cmark & \xmark & \cmark & \cmark & \xmark & $O(s \cdot f)$\\
\hline
 Bipartite BN   & Probabilistic  & IHU, IHU+ \cite{Steinder04} & \cmark & \xmark & \cmark & \xmark & \cmark & \cmark & \cmark & $O(s \cdot f^2)$\\
                        & Set covering          
                        &Max-covering, MCA+ \cite{Huang06} & \cmark & \xmark & \cmark & \xmark & \cmark & \cmark & \xmark & $O(s^2 \cdot f)$\\
                        & & SWPM \cite{Zhang08} & \cmark & \xmark & \cmark & \xmark & \cmark & \cmark & \xmark & $O(s \cdot f)$\\
                        & & HA, QOP \cite{Liao11} & $k$? & \xmark & \cmark & \cmark & \xmark & \cmark & \xmark & $O(s \cdot f)$\\
                        \cline{2-11}
                        & Marginals & MLA \cite{Chu09} & \cmark & \xmark & \cmark & \xmark & \cmark & \cmark & \cmark & $O(s \cdot f)$\\
                        \cline{2-11}
                        & MPE & Iterative MPE \cite{Steinder02} & \cmark & \cmark & \cmark & \xmark & \cmark & \cmark & \xmark &  $O(n^5)$\\
                        & &Positive Information \cite{Bouloutas94} & \cmark & \xmark & \cmark & \xmark & \cmark & \cmark & \xmark & $O(s \cdot f)$\\ 
                        & &PNIA \cite{Bouloutas94} & \cmark & \xmark & \cmark & \xmark & \cmark & \cmark & \xmark & NP-hard\\
                        & & GA \cite{Chu09} & \cmark & \xmark & \cmark & \xmark & \cmark & \cmark & \cmark & $O(s^2 \cdot f)$\\
                        && Probabilistic Multiple Fault \cite{Monacelli11} & $k$ & \cmark & \cmark & \xmark & \cmark & \cmark & \xmark & $O(s\cdot f\cdot f^k)$\\
\hline
Polytree BN & Marginals & Polytree\cite{Pearl89} & \cmark & \cmark & \cmark & \cmark & \cmark & \cmark & \xmark &  $O(n \ \mathrm{exp}(p))$\\
            \cline{2-11}
            & MPE & Iterative MPE \cite{Steinder02} & \cmark & \xmark & \cmark & \xmark & \cmark & \cmark & ? &  $O(n^6)$ \\
            
\hline
    General BN & Marginals (NP-hard) 
                        & Conditioning \cite{Pearl86} & \cmark & \cmark & \cmark & \xmark & \cmark & \cmark & \xmark & $O(n\exp(w))$\\
                        && Junction Tree \cite{Lauritzen90} & \cmark & \cmark & \cmark & \xmark & \cmark & \cmark & \xmark & $O(n\exp(w))$\\
                        &  & Arc reversal/Node elimination \cite{Shachter13} & \cmark & \cmark & \cmark & \xmark & \cmark & \cmark & \xmark& $O(n\exp(w))$\\
                        &  & Variable/Bucket elimination \cite{Zhang94}& \cmark & \cmark & \cmark & \xmark & \cmark & \cmark & \xmark & $O(n\exp(w))$\\
                        &  & Symbolic Inference \cite{Shachter90}& \cmark & \cmark & \cmark & \xmark & \cmark & \cmark & \xmark ? & $O(n\exp(w))$\\ 
                        &  & Differential Method \cite{Darwiche13}& \cmark & \cmark & \cmark & \xmark & \cmark & \cmark & \xmark & $O(n\exp(w))$\\
                        &  & Successive Restriction \cite{Smail05}& \cmark & \cmark & \cmark & \xmark & \cmark & \cmark & \xmark & $O(n\exp(w))$\\
                        & & Adaptive Inference \cite{Sumer11} & \cmark & \cmark & \cmark & \xmark & \cmark & \cmark & \cmark & $O(d^{3w}n \log n)$\\
                        &  & Adaptive Conditioning \cite{Ramos05} & \cmark & \cmark & \cmark & \xmark & \cmark & \cmark & \cmark & anytime\\
                        & & Search-based \cite{Poole96} & \cmark & \xmark & \cmark & \xmark & \cmark & \cmark & \xmark & $O(n\exp(w))$\\
                        & & Loopy Belief propagation \cite{Murphy99} & \cmark & \xmark & \cmark & \xmark & \cmark & \cmark & \xmark & $O(n\exp(w))$\\
                        & & Stochastic Sampling \cite{Geman84,Henrion86,Fung90,Shachter90b,Haque14}& \cmark & \xmark & \cmark & \xmark & \cmark & \cmark & \xmark & anytime\\
                        & & Markov Chain Monte Carlo \cite{Gilks95} & \cmark & \xmark & \cmark & \xmark & \cmark & \cmark & \xmark & anytime\\
                        \cline{2-11}
                        & MPE (NP-hard) & MPE bucket tree \cite{Dechter98} & \cmark & \cmark & \cmark & \xmark & \cmark & \cmark & \xmark & $O(n\exp(w))$ \\
                       
                       & & SLS \cite{Kask99}, WSAT \cite{Park02a} & \cmark & \xmark & \cmark & \xmark & \cmark & \cmark & \xmark & anytime \\
                       &  & MPE \cite{Kwisthout11} & \cmark & \cmark & \cmark & \xmark & \cmark & \cmark & \xmark & $O(n\exp(w))$ \\
                       &   & MPE (at most $k$ problems) \cite{Bahl07} & $k$ & \cmark & \cmark & \cmark & \xmark & \cmark & \xmark & $O(f^k)$\\
                       \cline{2-11}
                       & MAP (NP$^{\mathrm{PP}}$-hard) & MAP (Variable elimination, \ldots) \cite{Darwiche09} & \cmark & \cmark & \cmark & \xmark & \cmark & \cmark & \xmark & $O(n\  \exp(w_c))$\\
                       
                       &  & MAP (Adaptive Inference) \cite{Sumer11} & \cmark & \cmark & \cmark & \xmark & \cmark & \cmark & \cmark & $O(d^{3w}\log n + d^w m \log \frac{n}{m})$\\
                       & & Stochastic Sampling \cite{Geman84} & \cmark & \xmark & \cmark & \cmark & \cmark & \cmark & \xmark & anytime \\
                       & &MFE \cite{Kwisthout15}&\cmark & \cmark & \cmark & \cmark & \cmark & \cmark & \xmark&$\mathrm{NP}^{\mathrm{PP}^{\mathrm{PP}}}\!\mathrm{-hard}$\\
                       \cline{2-11}
                       & Probabilistic & Divide and conquer \cite{Katzela95} & \cmark & \xmark & \cmark & \xmark & \cmark & \cmark & \xmark & $O(n^3)$ \\
                       & Set covering &&&&&&&& \\
                       \cline{2-11}
                       & MAP simplification & eMAP \cite{Yuan07} & \cmark & \cmark & \cmark & \cmark 
                       & \cmark & \cmark & \xmark & ? ($O(2^f)$ inferences)
                       \\
                       & &MIE \cite{Kwisthout13}&\cmark & \cmark & \cmark & \cmark & \cmark & \cmark & \xmark&$\mathrm{NP}^{\mathrm{PP}}\!\mathrm{-hard}$\\
                       & &MRE exact \cite{Zhu15,Zhu16}&\cmark & \cmark & \cmark & \cmark & \cmark & \cmark & \xmark&$\mathrm{NP}^{\mathrm{PP}}\!\mathrm{-hard?}$\\
                       & &MRE local search \cite{Yuan09,Yuan11,Zhu15a}&\cmark & \xmark & \cmark & \cmark & \cmark & \cmark & \xmark&anytime\\
                       & &Explanation tree \cite{Flores05}&\cmark & \cmark & \cmark & \cmark & \cmark & \cmark & \xmark&$O(n^2\cdot d^2)$ inferences\\
                       & &Causal explanation tree \cite{Nielsen08}&\cmark & \cmark & \cmark & \cmark & \cmark & \cmark & \xmark&$O(n\cdot d)$ inferences\\

\end{tabular}\normalsize}
\end{center}
\end{table*}

In some cases Tables~\ref{table:inference:nonbayesian} and~\ref{table:inference:bayesian} only reflect some of the approaches for each type of technique, since, especially for Probabilistic Graphical Models, like Bayesian Networks, the amount of literature is very large. However, the general trend for dealing with complexity is common to all techniques: if the exact algorithm is too expensive (and they quickly become expensive for RCA as even the most basic problem setup based on the set covering algorithm in a bipartite non-probabilistic graph is already NP-hard), then compute an approximation. Approximate algorithms are usually inherently anytime, so they are a reasonable solution for real-time systems, at least if they are able to provide a sufficiently approximated answer in the provided time. For a good survey of anytime algorithms for Bayesian Networks refer to \cite{Ramos05}.

There is a large corpora of work related to resource constrained probabilistic inference\footnote{The term \emph{probabilistic inference} has been traditionally abused and can encompass the computation of marginals, MPE and MAP. However in many cases it just applies to the computation of marginals (i.e. posterior marginal probabilities of an arbitrary set of variables given an evidence)} in Bayesian Networks. For instance \cite{Guo02} reviews a number of strategies that can work real-time, broadly categorized between \emph{anytime algorithms} \cite{Garvey94}, which can provide an approximate answer at any time and converge to the correct solution as time goes by, and \emph{metalevel reasoners combined with multiple methods}. The class of anytime algorithms includes many algorithms for approximate inference, that start from a rough solution and iteratively approach the correct values. 

One big family of approximate methods are based on stochastic samplings, others are based on local search (e.g., MAP on Bayesian Networks has been approximated using genetic algorithms \cite{Campos99}, hill climbing and taboo search \cite{Park01} or simulated annealing \cite{Yuan04} to mention some) and a third class includes model simplification techniques \cite{Jensen90,Jaakkola99} that try to remove least significant elements (nodes/arcs/small probabilities) to obtain a smaller and more tractable system. Despite the approximations, often the resulting problem is still highly complex. 

In terms of complexity and performance, MPE and MAP, for general BNs, both exact and approximate, are NP-hard \cite{Shimony94,Abdelbar98}. In fact MAP complexity is $\textrm{NP}^{\textrm{PP}}$ complete \cite{Park02}, more specifically $O(n\  \exp(w_c))$, where $n$ is the number of variables and $w_c$ is the constrained tree width of the network, a value larger than the regular tree width $w$. On the other hand, MPE is $O(n\ \mathrm{exp}(w))$ \cite{Darwiche09}. For particular types of network topologies, like polytrees, these complexity results somewhat alleviate, as the tree width in a polytree is the maximum number of parents $p$ of any node, so MPE becomes $O(n \ \mathrm{exp}(p))$  but MAP is still NP-complete \cite{Park04}. Particular algorithms to compute the MPE are obviously satisfying these complexities: \emph{MPE bucket-tree} complexity is $O(\mathrm{exp}(n))$, while \emph{Iterative MPE} is $O(n^5)$ for bipartite graphs and $O(n^6)$ if used as an approximate algorithm for polytrees.


The approximate approach to compute MAP in \cite{Park01} simply turns it $O(n\  \mathrm{exp}(w))$, which is the same complexity as MPE or marginal computation. Nevertheless, exponential worse time computations does not imply that approaches are infeasible in practice. For instance the constraint-based search of \cite{Poole96} guarantees an error $\epsilon$ ($\epsilon < 0.5$) in the marginal computation for a fraction $1 - \Psi$ of all possible systems in time $O(n^c \log n)$ where $c\geq1$ is a constant that depends on $\epsilon$, $\Psi$ and $n$ among other factors \cite{Poole93}, and Loopy Belief Propagation can yield close approximations to the marginals in case it converges, each iteration cost being dependent on the diameter of the network, although the actual conditions that make it converge and be accurate are not generally understood \cite{Murphy99}.

Two other approaches by which MAP and MPE computation time can be reduced are either by using some precomputation strategy (like codebooks are precomputations of rules) or by assuming some fact that allows reducing the search space. For the first option \emph{Arithmetic Circuits} have been proposed (denoted as \emph{MPE (AC)} and \emph{MAP (AC)} in Table~\ref{table:inference:nonbayesian}). Their evaluation is linear on the circuit size, but they need to be compiled from the Bayesian Network and their size can be exponential in the number of variables (note that ACs can be learned directly, though). However, from a practical point of view they can reuse computations in the circuit, thus they are viewed as one of the fastest methods to compute MPE or MAP. For diagnosis they are good candidates, as they optimize the computation at the expense of having to recompile when the Bayesian Network changes its structure. An exponent of the second strategy, incorporating assumptions that allow reducing complexity, is \cite{Bahl07}, where they assume that faults are rare and, at any point, there will be at most $k$ simultaneous root causes of the observed symptoms (algorithm \emph{MPE (at most $k$ problems)} in Table~\ref{table:inference:bayesian}). In such a case, the complexity is $O(f^k)$ where $f$ is the number of possible root causes.

The easiest type of BN (besides the Na\"ive bayes classifier) is the bipartite BN, in which one set of nodes represents causes and the other nodes represent symptoms. The bipartite model is attractive because more complex propagation models can be reduced to this one (e.g., by graph reduction operations~\cite{Yemini96}) and it enjoys better computational tractability on approximate algorithms (exact solution is still exponential). However, the conversion from a more complex model to a bipartite model can bloat the final result: for instance in \cite{Steinder04} they model end-to-end failures in a network of size $n$ and the theoretical complexity of \emph{IHU} and \emph{IHU+} in that example is $n^4$. Although classified in the Bayesian Network approaches because it works on a graph annotated with probabilities \cite{Katzela95} can be seen as another example of this effect, although it never builds explicitly the bipartite graph, as it has cost $O(n^3)$.

The \emph{Positive Information} algorithm is conceptually similar to the \emph{Greedy set covering} as both try to select the minimum set of causes that would explain the symptoms, but using different although related heuristics. Since both algorithms are based on heuristics, they provide an approximation to the optimal solution in polynomial time. For \cite{Bouloutas94}, if probabilities involved in the computation do not have large differences, the solution is likely to be the optimal one. Algorithms \emph{IHU} and \emph{IHU+} work as well with this bipartite fault propagation model and have complexity $O(s \cdot f^2)$, where $s$ is the number of observed symptoms and $f$ is the number of potential faults considered, while algorithms \emph{Max-covering} and \emph{MCA+} have cost $O(s^2 \cdot f)$. The \emph{SWPM} algorithm has even a smaller cost $O(s \cdot f)$, akin of \emph{HA} and \emph{QOP} algorithms, which, besides using heuristics to reduce complexity, they also assume a small number of concurrent faults. 

We have seen in Section~\ref{sec:learn_rca_models} that decision tree models are fast to train. There are several approaches by which decision trees can be used for root cause analysis. One is to use them as classifiers by training them with observations labeled with a root cause. In such a case, the inference is simply done by traversing the decision tree until a leaf. This has a complexity $O(d)$, where $d$ is the size of the decision tree. There are several factors that affect the size of the tree generated, but in the worst case $d$ is $O(n)$, where $n$ is the number of training samples. However, such degenerated cases are rare, and on average $d$ is $O(\log_2 n)$.

Another possible use of decision trees is to distinguish only between observations with and without symptoms. In this case the decisions taken at each node of the tree leading to leafs with symptom observations are potential causes that could explain the symptoms. This is for instance the approach of the Four heuristics method \cite{Zheng04}, that applies four different heuristics to simplify the set of decisions. Although no complexity cost is detailed in \cite{Zheng04}, one of the steps involve merging decisions in the tree, thus we have assumed a simple search strategy which yields a complexity $O(d^2)$, but wiser implementations could have better cost. Along these lines, \cite{Chen04} mentions a greedy variant, deployed to diagnose at eBay, that only computes the decision on the decision tree that leads to the largest number of failures without actually building the tree. Since the implementation is not disclosed, the complexity is not available, although it is logical to assume that will be faster than \cite{Zheng04}. Finally, \cite{Aggarwal09} tries to find the ``easiest'' path (i.e., changes in the decisions taken in the tree) that could change from observations with symptoms to observations without symptoms. They consider that decisions have an associated cost, so they minimize this quantity, which could yield a solution with more changes than the fewest possible changes. In their case, they are able to automatically compute the costs by considering the frequency by which changes in decision naturally occur. Again, no specific implementation or complexity analysis is given in the paper and a na\"ive implementation by computing all alternative paths in the tree, computing its cost and then taking the minimum cost one would be $O(d^2)$. 

Most classifiers have good inference complexities. For instance the forward propagation in Neural nets highly depends on the architecture of the network, but as a rule of thumb, assuming an all connected network between layers, with $s$ input neurons (one per symptom), $f$ output neurons (one per fault), $h$ hidden neurons per layer and a total of $l$ layers, the number of operations is $O(\max(s,f,h)^2\cdot l)$. Most classifiers can be trained as single-class or multiple-class classifiers. In the table we assumed that the multiple-class neural network had the same structure as the single-class (for instance a threshold can be used to select the output labels), but other approaches are possible with different costs. On the other hand SVM classification cost is highly dependant on the type of kernel they use. For lineal SVMs, the cost is linear in $s$, assuming the symptoms are the features used. For kernel SVMs, it can be $O(v\cdot s)$ considering there are $v$ support vectors if the kernel is linear on the number of features, however more complex kernels will increase that cost. 

Not surprisingly, other models that have a reasonably fast inference are the ones based on codebooks, as they can be viewed as a type of nearest neighbor classifier. For instance the complexity of \cite{Yemini96} is the complexity of a minimum Hamming distance decoding. There are several possible implementations ranging from exhaustive comparison with each symptom vector for each fault, with the cost $O(f \cdot s)$ that appears on the table, to Standard Array or Syndrome decoding that are faster but require more space (so they have a larger learning cost). The disadvantage of codebooks is that they consider only a single-failure model. The technique in \cite{Monacelli11} extended it to multiple faults by allowing considering up to $k$ simultaneous faults, by constructing a codebook with all the possible combinations of faults, thus the cost of inference is the one proportional to a codebook of $f^k$ entries that, as we have seen before, depends on the actual implementation of the minimum Hamming distance decoding. The table reports the cost for the basic exhaustive search strategy ($O(f^k \cdot s)$).

Results on logics are as usual in consonance with their expressiveness: logic abduction in first-order logic is undecidable because the set of explanations may be infinite, thus deciding if an explanation is minimal is generally undecidable \cite{Mayer93}. Logic abduction on propositional logic is decidable, but its complexity is large (deciding if there is an explanation is more complex than NP-complete) and depends on the types of clauses allowed (e.g., Horn clauses are easier than the general proposition logic), see \cite{Eiter95} for a detailed explanation and \cite{Nordh08} for a summary of approaches to tackle that complexity, including heuristics, reduction to QBF, compilation and approximations.

Dempster-Shafer theory allows associating degrees of belief and plausability to logical propositions. These propositions can be combined using the Dempster-Shafer combination rule. However the application of this rule is $\#P-complete$ \cite{Orponen90}, thus the straightforward implementation of the belief update algorithm \cite{shafer76} is exponential on the size of the propositions. Consequently, approximate ways to compute the resulting beliefs have been proposed \cite{Wilson91}. For a comprehensive review of alternative approaches and complexities see \cite{Wilson00}.

An interesting family of models that have tractable inference are Sum-Product Networks (closely related to Arithmetic Circuits), that have inference algorithms to compute marginals or MPE linear in the size of the model. Relational Sum-Product Networks do not have specific inference algorithms but just work as templates to generate SPNs and then apply the inference algorithms for them.

Markov Logic Networks are an alternative formalism to Bayesian Networks. By default they support deduction, but adding some logical formulas it is possible to make them perform abduction as well, as shown by \cite{kate09} and \cite{singla11}. These two techniques differ in the amount of nodes they add to the MLN and how they impact the treewidth of the graph, to whom there is an exponential dependency in terms of complexity. Marginal computation can be done using lifted inference \cite{Braz07,Broeck11}. In the latter case they do it by transforming the MLN into a first order deterministic decomposable negation normal form, which has an inference complexity polynomial on the size of the formula. Finally \cite{Kazemi16} takes the approach from \cite{Broeck11} but transforms the inference problem into C++ code that can be compiled, thus has the same asymptotic complexity but can yield notable accelerations from a practical point of view.

The \emph{Most frugal explanation}\cite{Kwisthout15} is a heuristic approach to compute MAP. Although its complexity is $\mathrm{NP}^{\mathrm{PP}^{\mathrm{PP}}}\!\mathrm{-hard}$, thus, more complex than MAP itself, it becomes tractable under a set of constraints that is less strict than the one that makes MAP tractable.

The \emph{Divide and conquer} algorithm is a two-phase algorithm in which first the nodes of the graph are aggregated by maximum mutual dependency and then a subset is selected that explains all symptoms and contains a fault. Its complexity is $O(n^3)$, where $n$ is the number of nodes in the graph.

It is possible to diagnose not only snapshots of symptoms but as well when symptoms appear as sequences or processes. For instance the approach in \cite{Aghasaryan98} uses partially stochastic Petri nets to find the most likely sequence of transitions (faults) that happened in the system to produce the observed sequence of symptoms.

Besides Marginal, MPE and MAP inference concepts, a number of extensions have been proposed, since in many cases they can produce overspecified explanations \cite{Pacer13}. For instance \emph{explanatory MAP} \cite{Yuan07} identifies the most relevant variables from a subset of the non-observed variables, where the relevancy can be a metric like the Bayes factor or the Likelihood \cite{Yuan07}. \cite{Zhu15,Zhu16}  follow a similar approach but using a different metric based on the generalized Bayes factor (they compute the ratio between the probability of a particular hypothesis over the probability of the rest of hypotheses). In these cases we have marked these approaches as able to work with unknown causes, as the whole set of non-observed variables could be used as the target set, effectively letting the task of deciding what variables had greater effect on the observations to the algorithm. The search space for MRE is very large, thus exact algorithms have conjectured complexity $\mathrm{NP}^{\mathrm{PP}}\!\mathrm{-hard}$ \cite{Yuan11} and several local search algorithms have been developed \cite{Yuan09,Yuan11,Zhu15a}.

On the other hand the \emph{Most Inforbable Explanation} \cite{Kwisthout13} tries to introduce a balance between how informative should be an explanation and how probable it is, two requirements that are generally opposed: a more specific information (thus less probable than a generic one) is considered more informative than a general one. Computing MIE has complexity $\mathrm{NP}^{\mathrm{PP}}\!\mathrm{-hard}$ although it becomes tractable under a similar set of constraints that render MAP tractable.

Besides these algorithms that provide an assignment of a subset of variables as an output, tree-based approaches have been investigated as well \cite{Flores05,Nielsen08}. The idea behind these models is that if the number of variables in the subset is very large, users should have a help to understand the set, by creating a tree in which every node is a variable and sorting them by relevance. At each step, the tree is grown considering the evidence plus the hypothesis previously seen in that branch of the tree. The difference between the methods is the metric used to choose the best variable to add, how
they decide that growth should stop, and how they then evaluate the resulting hypotheses. 

Some of these extended explanatory models beyond MAP, that we have collectively labeled as \emph{MAP simplification} techniques do not have a complexity cost specified as in the previous techniques. Rather they have to evaluate the a posteriori probability of some hypothesis given some evidence at each step of their computation. Thus the complexity we give in some cases is in the number of calls to an underlying computational model (a BN, for instance) used to compute these posterior marginals (denoted by the suffix \emph{inferences} in the table). As we have seen, for many models, the complexity of such a computation might be prohibitive in the general case, a fact that might partially explain why these more advanced explanatory models had not a more widespread adoption.

The complexity of Causal explanation trees \cite{Nielsen08}, in terms of number of calls to an inference engine per node in the constructed tree, is $O(n d)$, where $n$ is the number of explanatory variables and $d$ is the average domain size of the variables, e.g., 2 for binary variables. Explanation trees \cite{Flores05} approach is $O(n^2 d^2)$.



Finally, one way to cope with the complexity of the inference task is to parallelize the algorithms. The survey in \cite{Alves15} covers parallel inference in Bayesian Networks, most of them parallelizing the \emph{Junction Tree} algorithm. Approaches ranged from multi/many-core algorithms based on OpenMP \cite{Namasivayam06}, with cost $O(n^2w/p+wd^wn/p+n \log p)$, where $n$ is the number of nodes in the Bayesian network, $w$ is the clique width, $d$ is the number of possible states of the variables and $p$ is the number of processors, and scalable over $1 \leq p \leq nw/ \log n$, to multi-core with pthreads \cite{Xia11}, GPUs \cite{Zheng12}, hybrid CPU-GPGPU systems \cite{Jeon10} or FPGAs \cite{Lin10}.

\section{Conclusion}
\label{sec:conclusion}

In this survey we have reviewed the models and algorithms to perform inference that have been used for Root-Cause analysis. The survey shows a wide spectrum of techniques with the usual trade-off between tractability and expressiveness. As future work we plan to extend this survey to consider specifically RCA for Big Data, by providing a users' guide to RCA in this setting.

\section*{Acknowledgment}

The authors would like to thank LeanBigData (FP7-619606) project.


\bibliographystyle{IEEEtran}
\bibliography{anomalydetection}

\end{document}